\documentclass[final]{cvpr}

\usepackage{float}
\usepackage{times}
\usepackage{epsfig}
\usepackage{graphicx}
\usepackage{amsmath}
\usepackage{amssymb}

\usepackage{multirow}
\usepackage{tabularx}
\usepackage{booktabs}
\usepackage{amssymb}
\usepackage{pifont}

\usepackage{setspace}
\usepackage{graphicx}
\usepackage{colortbl}

\usepackage[dvipsnames]{xcolor}

\newcommand{\cmark}{\ding{51}}%
\newcommand{\xmark}{\ding{55}}%

\usepackage{algorithm} 
\usepackage{algorithmic}

\makeatletter\renewcommand\paragraph{\@startsection{paragraph}{4}{\z@}
  {.5em \@plus1ex \@minus.2ex}{-.5em}{\normalfont\normalsize\bfseries}}\makeatother

\def\eg{\textit{e.g.}}
\def\ie{\textit{i.e.}}

\def\etal{\textit{et al.}}

\usepackage[pagebackref=true,breaklinks=true,colorlinks,bookmarks=false]{hyperref}

\def\cvprPaperID{508} 
\def\confYear{CVPR 2021}

\begin{document}

\title{Boundary-sensitive Pre-training for Temporal Localization in Videos}

\author{Mengmeng Xu$^{1,2}\thanks{Work done during an internship at Samsung AI Centre.}$
\and
Juan-Manuel P\'erez-R\'ua$^{1}$
\and
Victor Escorcia$^{1}$
\and
Brais Martínez$^{1}$
\and
Xiatian Zhu$^{1}$
\and
Li Zhang$^{1}$
\and
Bernard Ghanem$^{2}$
\and
Tao Xiang$^{1,3}$
\and
{\small $^1$ Samsung AI Centre Cambridge, UK}\\
{\tt\small \{j.perez-rua,v.castillo,brais.a,xiatian.zhu,li.zhang1,tao.xiang\}@samsung.com}\\
{\small $^2$ King Abdullah University of Science and Technology (KAUST), Saudi Arabia}\\
{\tt\small \{mengmeng.xu,bernard.ghanem\}@kaust.edu.sa}\\
{\small $^3$ University of Surrey, UK}
}

\maketitle

\begin{abstract}
Many video analysis tasks require temporal localization for the detection of content changes. However, most existing models developed for these tasks are pre-trained on general video action classification tasks. This is due to large scale annotation of temporal boundaries in untrimmed videos being expensive. Therefore, no suitable datasets exist that enable pre-training in a manner sensitive to temporal boundaries. In this paper for the first time, we investigate model pre-training for temporal localization by introducing a novel boundary-sensitive pretext (BSP) task. Instead of relying on costly manual annotations of temporal boundaries, we propose to synthesize temporal boundaries in existing video action classification datasets. By defining different ways of synthesizing boundaries, BSP can then be simply conducted in a self-supervised manner via the classification of the boundary types. This enables the learning of video representations that are much more transferable to downstream  temporal localization tasks. Extensive experiments show that the proposed BSP is superior and complementary to the existing action classification-based pre-training counterpart, and achieves new state-of-the-art performance on several temporal localization tasks.
\end{abstract}
\begin{figure}[t]
    \centering
    \includegraphics[trim={6cm 0cm 6cm 0cm},width=8.5cm,clip]{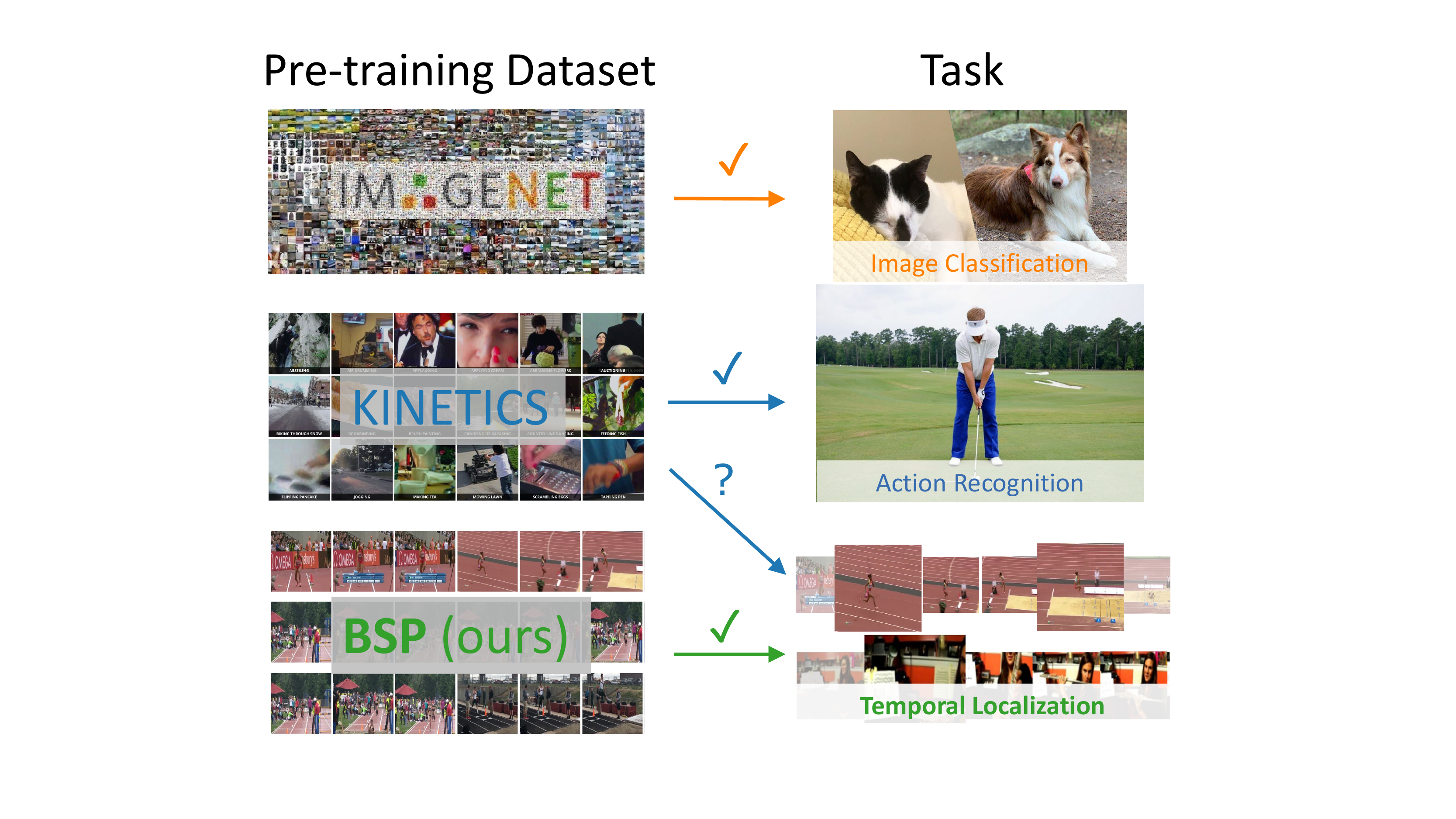}
    \vspace{-8mm}
    \caption{
    \textbf{Pre-training datasets for different tasks.}
    The well-established pre-training-then-fine-tuning paradigm for image and video classification model optimization is effective thanks to the availability of large related datasets (\eg, ImageNet and Kinetics).
    For video temporal localization tasks however, existing datasets are either too small for model pre-training or less effective due to the lack of temporal boundary annotations.
    We solve this problem by introducing 
    a novel boundary-sensitive pretext (BSP) task.
    } 
    \vspace{-3mm}
    \label{fig:intro}
\end{figure}

\section{Introduction}

Recently, the focus on video analysis has shifted beyond trimmed video action classification to video temporal localization. This is because in many real-world applications, instead of short (\eg, few seconds long) video clips,
{\em long, untrimmed} videos are often presented (\eg, from social media websites as YouTube, Instagram) with both non-interesting background and foreground contained \eg, a particular action of interest. This requires a video model to conduct {\em temporal localization tasks}. Examples of these tasks include 
temporal action localization \cite{xu2020g,Buch2017SSTST},
video grounding \cite{Hendricks_2017_ICCV,Mun_2020_CVPR}
and step localization~\cite{Zhukov2019}.

As in most other visual recognition tasks, recent models designed for video temporal localization are based on deep learning. As such, model pre-training is critical. 
In particular, a
two-staged model training strategy is commonly adopted \cite{han2020self,wang2020self,han2020memory,benaim2020speednet}: first, a video encoder is pre-trained on a large action classification dataset (\eg, Kinetics \cite{carreira2017quo}\footnote{Kinetics was created by trimming 10s clips around a frame manually labelled as containing the target action. Therefore the rest
of the untrimmed video cannot be treated as background.}, Sports-1M \cite{karpathy2014large}), then, a temporal localization head is trained on the target small-scale temporal localization dataset leaving the video encoder fixed. Thus, there is a clear mismatch between the pre-training of the video encoder and the target task. 
Ideally, model pre-training should be carried out on temporal boundary-sensitive tasks. However, this is not possible due to the lack of large scale video datasets with temporal boundary annotation. This is because temporal boundaries' labeling is
much more expensive and tedious than video-level class labeling due to the need for
manually examining every single frame\footnote{Boundary annotations are $~3.8\times$ more expensive than class annotations \cite{caba2015activitynet,zhao2019hacs}}.
%
%
%

In this paper, we investigate the under-studied yet critical problem of {\em model pre-training for temporal localization in videos}. Due to the difficulties in collecting a large-scale video dataset with temporal boundary annotations, 
we propose to synthesize large-scale untrimmed videos with temporal boundary annotations by transforming the existing trimmed video action classification datasets. Once the pre-training data problem is tackled, we focus on defining and evaluating a number of pretext tasks capable of exploiting the particularities of the synthesized data through self-supervision.

In particular, the first key challenge boils down to how to obtain large scale training video data with temporal boundary information in a scalable and cheap manner. To that end, we introduce a simple yet effective method for generating three types of temporal boundary at large scale 
using existing action classification video data (\eg Kinetics). More specifically, we generate artificial temporal boundaries corresponding to video content changes by either stitching trimmed videos containing different classes, stitching two video isntances of the same class, or by manipulating the speed of different parts of a video instance. 
The associated pretext task used to train the video model uses standard supervised classification learning, where the task is to distinguish between the types of temporal boundaries as defined above. We experimentally show that such task offers superior performance to other possible pretext tasks, such as regressing the temporal boundary location, and that combining the different boundary types into a multi-class classification problem is superior to all binary classification tasks in isolation.

%

The following {\bf contributions} are made in this work:
(I) We investigate the problem of model pre-training for temporal localization tasks in videos,
which is largely under-studied yet particularly significant to video analysis.
(II) We propose a scalable video synthesis method that can generate
a large number of videos with temporal boundary information.
This approach not only solves the key challenge of lacking large pre-training data, but also facilitates the design of model pre-training. 
%
%
(III) 
Extensive experiments show that temporal action localization, video grounding, and step localization tasks can significantly benefit from the proposed model pre-training,
yielding compelling or new state-of-the-art performance on a number of benchmark datasets.

\section{Related work}\label{sec:related}
\paragraph{Temporal localization tasks.}
Temporal localization in videos encompasses tasks such as temporal action localization (TAL), video grounding and step localization.
Although those tasks have their own particularities, they share the same target: recognizing the specific point in time where the semantic content of the video changes. TAL focuses on predicting the temporal boundaries and class of an action instance in untrimmed videos \cite{caba2015activitynet}.
Instead, video grounding generalizes temporal action localization by not relying on a predefined set of action categories \cite{Hendricks_2017_ICCV,gao2017tall,escorcia2019temporal}, the task being to localize the segment in the video that best matches a given language query.
Step localization~\cite{Zhukov2019} is associated with the detection of different actions involved in the execution of a complex task, \eg~\textit{Change tire}, in instructional videos~\cite{miech19howto100m}. Instructional videos are highly edited audio-visual tutorials with aesthetical transitions and cuts.

Due to constraints on computational cost, solutions to these tasks are usually not based on end-to-end training. Instead, a video encoding network is pre-trained on an action classification task to enable large-scale training. 
A temporal localization model is then trained using a fixed feature extraction backbone.
%

\paragraph{Video encoding networks.}
It is common among state of the art methods to use a pre-trained network as the video encoder. Such network is trained on a \textit{classification} task using standard cross-entropy, typically on large-scale datasets such as Kinetics~\cite{zisserman2017kinetics,Zeng_2020_CVPR}. For example, it is common for TAL, \eg, \cite{lin2018bsn,lin2019bmn,xu2020g,Bai2020bcgnn}, to use features extracted with a two-stream \cite{simonyan2014two} TSN model~\cite{wang2016temporal,trn_eccv18}. That is, the model comprises two TSN networks, one with ResNet50~\cite{he2016deep} backbone trained on RGB, and another with a BN-Inception backbone~\cite{ioffe2015batch} trained on Optical Flow. Other methods use 3D CNN-based models such as two-stream I3D models~\cite{carreira2017quo}, \eg, \cite{he2019rethinking,Zeng2019GraphCN}, or Pseudo-3D~\cite{qiu2017learning}, \eg, \cite{Long2019GaussianTA}.
Alternatively, some methods exploit the temporal segment annotations on the downstream temporal localization datasets to define a classification task, and use it to pre-train the video encoder~\cite{lin2019bmn,xu2020g,Rodriguez_2020_WACV,Mun_2020_CVPR}. This results in less of a domain gap, at the cost of large-scale training.


A few methods further add an end-to-end fine-tuning stage directly on the downstream tasks, e.g. R-C3D~\cite{xu2017r}, PBR-net~\cite{liu2020progressive}. However,  end-to-end training is achieved by compromising other important aspects, e.g. using batch size of 1, resulting in lower performance in practice. 

Although an action classifier trained through cross-entropy can represent the overall content of a video segment, a feature extractor trained in this manner is not tuned to be sensitive to specific temporally-localized structures, such as the start or end of an action. Instead, we propose a boundary-sensitive self-supervised pre-training that results in features with the desired temporally-localized sensitivity.

\paragraph{Temporal localization heads.}

\textit{Temporal action localization} methods follow either a two-stage or a one-stage approach. Two-stage methods first generate candidate action segments (e.g., proposals)~\cite{Buch2017SSTST, Heilbron2016FastTA, Escorcia2016DAPsDA, Liu2019MultiGranularityGF, Gao2018CTAPCT}, and then use a classifier on each proposal to obtain a class score~\cite{Shou2016TemporalAL, shou2017cdc, Zeng2019GraphCN, Zhao2017TemporalAD, lin2018bsn}. One-stage methods predict instead the temporal action boundaries or generate the proposals, and classify them in a shared network~\cite{heilbron2017scc, chao2018rethinking, xu2017r, Yuan2017TemporalAL, xu2020g, Long2019GaussianTA, lin2019bmn, Bai2020bcgnn}.

\textit{Video grounding} is similar to temporal action localization, but requires a language model. The current literature can also be clustered into the two groups. (1) Proposal-based approaches adopt a propose-and-rank pipeline~\cite{chenhierarchical,ghosh_etal_2019_excl, Liu_2018_ECCV,GDP_2020_AAAI,lu_etal_2019_debug, chen2020learning,yuan2019to,zhang_etal_2020_span}, relying first on a proposal model much like for temporal action localization, and then rank the resulting snippets based on their similarity with the textual query~\cite{Ge_2019_WACV, song2018val,chen_etal_2018_temporally, wang2020temporally,YuDLZ00L18,zhang2019man}. 
(2) Proposal-free methods~\cite{Hendricks_2017_ICCV, liu2018crossmodal, wu2018multi,Mun_2020_CVPR} directly regress the temporal boundaries of the queried moment from the multi-modal fused feature information. 


\textit{Step localization in instructional videos}. The task corresponds to the alignment of a set of steps required to complete a task, in the form of text entries, and a video exemplifying such task~\cite{miech19howto100m}. Recently,~\cite{Zhukov2020} showed improved  step localization performance over multiple models when using an actionness-based proposal generation method.


In order to show the benefit of our proposed boundary-sensitive pre-training, we adopt a set of recent publicly-available methods representative of the current state of the art for these tasks: G-TAD~\cite{xu2020g}, LGI~\cite{Mun_2020_CVPR} and 2D-TAN~\cite{2DTAN_2020_AAAI}. In order to keep a fair comparison with prior work, we do not fine-tune the video encoding network on the downstream dataset. When use our BSP features in conjunction with these models, we use the default training configurations defined by the respective authors, and report the performance using the provided evaluation scripts.

\paragraph{Self-supervised learning in videos.}

While current temporal localization literature focuses on pre-training through supervised learning, the rapid advancement of self-supervised learning makes it a promising alternative to sidestep end-to-end training \cite{alwassel2020xdc,benaim2020speednet,miech20endtoend,shuffle_learn_eccv16,arrow_cvpr18}. 
Among these, a large body of research has focused on finding effective temporal-related pretext tasks. Some works considered frame ordering, either learning through triplets of frames \cite{shuffle_learn_eccv16}, through sorting a sequence \cite{sorting_seq_iccv17}, or by distinguishing whether sequences are played forward or backwards \cite{arrow_cvpr18}. Alternatively, pretext tasks related to the speed of the video have recently become popular~\cite{benaim2020speednet,playback_rate_cvpr20,wang2020self}. 
An effective variant on this theme was proposed in \cite{temporal_transform_eccv20}, where 
clips undergo one among a set of possible augmentations, and at the same time are also sampled at one of a set of possible frame-rates. The pretext task is then to correctly classify both the playback speed and the temporal augmentation applied. 
Alternative approaches include predicting motion-related statistics~\cite{stats_cvpr19} and more direct extensions of successful image-based contrastive learning methods to the video realm~\cite{st_contrastive_arxiv20}.

These methods exploit video-specific characteristics to force the network to focus on the semantic content within the video, inducing representations capturing long-term temporal semantic relations, but forces invariance to the relative positioning of the sampled segment within the action. They thus cannot be adopted for pre-training a temporal localization model. 

\newcommand*{\isotrib}[1]{%
  {\color{cyan}\scalebox{1.118034}[1]{$#1$}}%
}
\newcommand*{\isotrir}[1]{%
  {\color{red}\scalebox{1.118034}[1]{$#1$}}%
}
\newcommand*{\isotrig}[1]{%
  {\color{YellowGreen}\scalebox{1.118034}[1]{$#1$}}%
}

\begin{figure*}[t]
	\begin{center}
		\setlength{\tabcolsep}{0.25mm}
		\renewcommand{\arraystretch}{1.5}
		\begin{tabular}{ccc}
		
        \rotatebox{90}{\parbox{4cm}{\centering \footnotesize Diff-class boundary}} &
        \rotatebox{90}{\parbox{4cm}{\centering \scriptsize \hspace{-3mm}Synthesized\hspace{2mm}Kinetics\hspace{3mm}ActivityNet}} &              
        \includegraphics[width=0.96\linewidth]{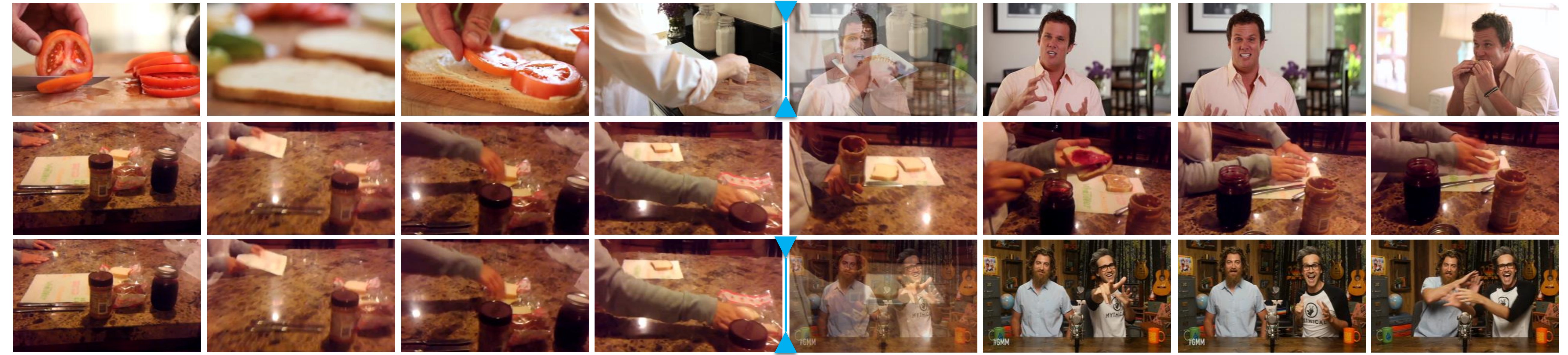} \\[-4mm]
		
        \rotatebox{90}{\parbox{4cm}{\centering \footnotesize Same-class boundary}} &
        \rotatebox{90}{\parbox{4cm}{\centering \scriptsize \hspace{-3mm}Synthesized\hspace{2mm}Kinetics\hspace{3mm}ActivityNet}} &              
        \includegraphics[width=0.96\linewidth]{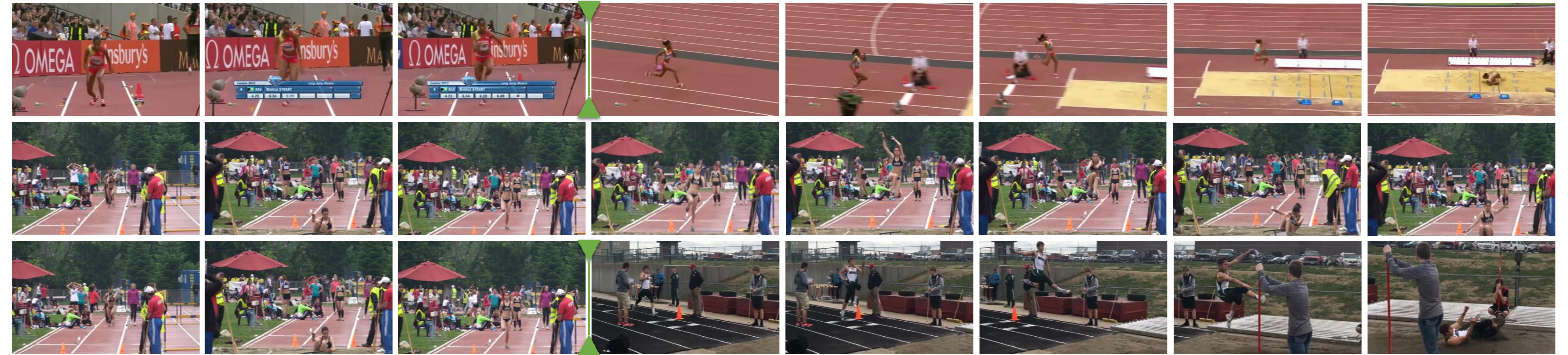} \\[-4mm]
        
        \hspace{-1mm}\rotatebox{90}{\parbox{4cm}{\centering \footnotesize Diff-speed boundary}} &
        \rotatebox{90}{\parbox{4cm}{\centering \scriptsize \hspace{-3mm}Synthesized\hspace{2mm}Kinetics\hspace{3mm}ActivityNet}} &              
        \includegraphics[width=0.96\linewidth]{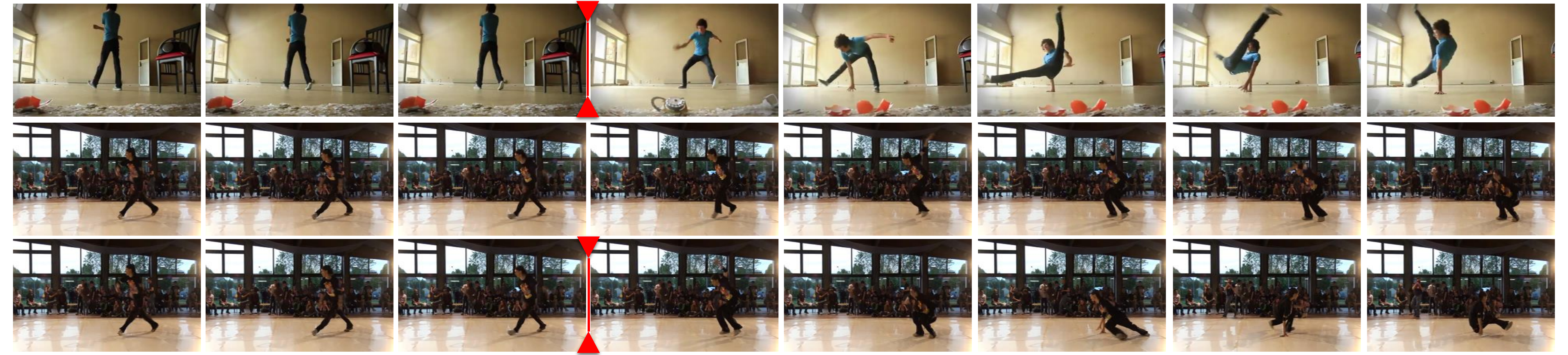} 
		\end{tabular}
	\end{center}
    \vspace{-3mm}
    \caption{
     \textbf{Illustration of our {\em boundary-sensitive video} synthesis.}
     For each group of three rows we show from top to bottom: 
     (a) a \textbf{real clip} from ActivityNet with real action class boundary; (b) a clip from Kinetics-400 with no boundaries; and (c) a \textbf{synthesized video} by one of the proposed methods using samples from Kinetics-400 with synthetic boundaries. 
    \textbf{(Top)} Diff-class boundary (\isotrib\blacktriangledown); two clips from different categories are smoothly merged around frame \#5.     
    \textbf{(Middle)} Same-class boundary (\isotrig\blacktriangledown); two clips from the same Kinetics category are stitched together from frame \#4.
    \textbf{(Bottom)} Diff-speed boundary (\isotrir\blacktriangledown); a clip from Kinetics is artificially sped up from frame \#4.
    }
    \vspace{-4mm}
    \label{fig:synthesis}
\end{figure*}

\section{Method}

\subsection{Problem context}
We consider the model pre-training problem 
for temporal localization in videos.
This is the first stage of the common {\em first pre-train backbone, then train localization head} paradigm, and the downstream tasks in this context
include temporal action localization \cite{xu2020g}, 
video grounding~\cite{Hendricks_2017_ICCV}, and 
step localization~\cite{Zhukov2019,Zhukov2020}.
The {\em vanilla pre-training} method simply conducts
supervised learning on a large video dataset (\eg, Kinetics)
$D_{tr} = \{\textbf{V}_i\}_{i=1}^N$ with action class labels as ground-truth supervision.
This brings about action content awareness to the model.
As a result, such a pretrained model is usually superior 
to those with random initialization and pretrained on 
image data (\eg, ImageNet). However, this method is limited 
in capturing {\em temporal boundary information} as required 
by temporal localization tasks, because
boundary annotations are not available in existing 
large video datasets.

To overcome the above limitation of action classification-based pre-training,
we aim to directly address the unavailability of boundary labels
in videos, which is intuitive though non-trivial. 
As a large number of video data is needed for pre-training,
the method must be {\em scalable} without expensive manual annotation.
We therefore adopt a data synthesis approach.
Existing trimmed video datasets are selected as the video source
since they exist in large scale.
In our implementation, Kinetics \cite{carreira2017quo} 
is chosen as it is the most common selection 
for vanilla pre-training of a video encoder model in the literature.
Moreover, this selection allows to avoid extra training data
which further ensures accurate evaluation and fair comparisons.

Given trimmed video data with action class labels,
we introduce four temporal boundary concepts including 
{\em different-class boundary, same-class boundary, different-speed boundary, same-speed boundary}.
They all require {\em {zero}} extra annotation in video synthesis
and hence enable us to generate an arbitrary number of video samples with boundary labels. 
Next, we will 
describe the proposed {\em boundary-sensitive video synthesis} method.


\subsection{Boundary-sensitive video synthesis}
\label{sec:synthesis}
{Temporal boundary refers to a transition of shots or scenes, or a change of action content.}
In this work, we consider two perspectives of the video source:
class semantics and motion speed, both of which are available in Kinetics (i.e., class label and frame rate).
Four different boundary classes are then formulated as detailed below.

\paragraph{(1) Diff-class boundary.}
This boundary is defined as the edge
between two action instances from {\em different classes}.
It is the most intuitive boundary, as typically presented
in untrimmed videos with different actions taking place continuously.
To synthesize a video with this boundary, 
we use two videos $\textbf{V}_1$ and
$\textbf{V}_2$ sampled randomly from different action categories.
Specifically, we first perform uniform frame sampling in each video, 
$F_1=\{f_{1, i}\}_{i=1}^{\tau + \epsilon} \subset \textbf{V}_1$ and
$F_2=\{f_{2, i}\}_{i=1}^{\tau + \epsilon} \subset \textbf{V}_2$,  
where $f_{1,i}$ ($f_{2,i}$) denotes the $i$-th frame from $\textbf{V}_1$ ($\textbf{V}_2$), and
$\tau + \epsilon$ frames are sampled from each video.
Then a new video $S_{dc} = \{f_i^{dc}\}_{i=1}^{2\tau} $ 
(Fig.~\ref{fig:synthesis}(a))
is synthesized with each frame formed as:
%
\begin{align} \footnotesize
f_i^{dc} =
    \begin{cases}
    f_{1,i} & {i \in [1, \tau-\epsilon]}, \\
    \omega_1(i) f_{1,i} + \omega_2(i) f_{2, i-\tau+\epsilon} 
    & {i \in (\tau-\epsilon, \tau+\epsilon] }, \\
    f_{2,i-\tau+\epsilon} & {i \in (\tau+\epsilon, 2\tau]}.
    \end{cases}
\label{eq:trans}
\end{align}
where $\epsilon$ controls the period of action transition.
The transition is made by a weighted blending approach with the weights defined as
$\omega_1(i) =\frac{1}{2\epsilon}(\tau+\epsilon-i)$, $\omega_2(i) = \frac{1}{2\epsilon}(i-\tau+\epsilon)$.
This transition injects some smoothing effect from one action to the other. As a result, the output video $S_{dc}$ 
eliminates abrupt content change, making the following model
pre-training meaningful without trivial solution.




\paragraph{(2) Same-class boundary. } 
Complementary to diff-class boundary, 
this aims to simulate the scenarios 
where the same action happens repeatedly and continuously.
This is frequently observed in untrimmed videos 
with multiple different shots of the same action class
in a row. 
Similarly, we select two videos $\textbf{V}_1$ and
$\textbf{V}_2$ from the {\em same} action class
and sample $\tau$ frames $F_1 =\{f_{1, i}\}_{i=1}^{\tau}$ and 
$F_2 =\{f_{2, i}\}_{i=1}^{\tau}$ from each selected video.
A new video $S_{sc}=\{f_{i}^{sc}\}_{i=1}^{2\tau}$ 
(Fig.~\ref{fig:synthesis}(b))
is synthesized by concatenation as:
\begin{align}
f_i^{sc} =
    \begin{cases}
    f_{1,i} & {i \in [1, \tau]}, \\
    f_{2,i-\tau} & {i \in (\tau, 2\tau]}.
    \end{cases}
\label{eq:trans}
\end{align}
Transition is not applied in this case,
since the semantic content is similar in the two input videos.





\paragraph{(3) Diff-speed boundary. }
This boundary class is motivated by an observation that
the speed of content change varies from background (\eg, without action) to foreground (\eg, with action) and from one action instance to the other.
Hence speed change entraps potentially useful temporal boundary information.
Formally, we start with sampling a random video $\textbf{V} = \{f_1, f_2,\dots\}$ from the source data. 
Then, a new video $S_{ds} = \{f_i^{ds}\}^{2\tau}_{i=1}$ with two 
different speed rates (Fig.~\ref{fig:synthesis}(c)) 
is generated by
\begin{align}
    f_i^{ds}
    & =
    \begin{cases}
    f_{i} & {i \le t} \;\; (\text{original rate}), \\
    f_{t+\gamma(i-t)} & {i > t} \;\; (\text{new rate}).
    \end{cases}
\label{eq:acc}
\end{align}
where $t$ is the change point of speed 
and $\gamma \neq 1$ denotes the sampling rate introduced in video synthesis.
When $\gamma \in (0, 1)$, temporal upsampling
is triggered. 
When $\gamma > 1$, temporal downsampling takes place. 
By varying $\gamma$, a large number of speed change cases can be simulated.  
In case the frame index ${t+\gamma(i-t)}$ is not an integer,
we simply use the nearest frame.
More complex frame interpolation can be considered
which however was found to have no clear advantage in performance.


\paragraph{(4) Same-speed boundary. }
This is introduced to serve as a {\em non-boundary} class
for conceptual completion.
For this class, the same videos from the source set are 
used with the coherent original speed rate throughout all the frames in each video. 
For notation consistency, we denote the videos of this type as
$S_{ss}$.

Collectively, we denote all four types of boundary-sensitive videos as $S = \{S_{dc}, S_{sc}, S_{ds}, S_{ss} \}$.

\subsection{Boundary-sensitive pre-training}
\label{ssec:pretexts}

Given the boundary-sensitive video data $S$ as generated in
Sec.~\ref{sec:synthesis}, we now describe how to use them for
video model pre-training so that the pre-trained model
can benefit the temporal localization downstream tasks.
For simplicity and easy adoption of our method,
we consider two common supervised learning algorithms
based on the synthetic boundary information.

{\bf Pre-training by classification ({\em default choice}).}
An intuitive pre-training method is supervised classification
by treating each type of synthetic video as a unique class.
That is, a four-way classification task.
Formally, we first label a boundary class $y\in\{0,1,2,3\}$
to each video $x \in {S}$ according to their boundary type.
With a target video model $\theta$,
we predict the boundary classification vector $\textbf{p}=\{p_0, p_1, p_2, p_3\}$ for a given training video $x$.
To pre-train the model, we use the cross-entropy loss function.
The main merit of this method is to easily accommodate different types of boundary supervision in a principled manner.
This allows the effective use of our synthetic video data.

{\bf Pre-training by regression. }
An alternative way for pre-training with our training data
is the change point regression.
For more stable learning, we convert the ground-truth change point $\mu$ 
into a 1D Gaussian heatmap $\textbf{y} = [y_1, \cdots, y_t, \cdots, y_L]$ as
the regression target, $y_t = \exp{[-\frac{1}{2\tau}(t-\mu)^2]}$ for $t \in [1, L]$,
where we sample a snippet comprising $\tau$ frames from the video with length $L$.
Let the model prediction output be $\textbf{r} = [r_1, \cdots, r_t, \cdots, r_L]$.
We minimize the smooth $\mathcal{L}_1$ norm of $(\textbf{y}-\textbf{r})$.


\begin{figure}[t]
         \hspace{-2mm}\includegraphics[trim={4cm 1cm 3cm 4.5cm},width=9.0cm,clip] {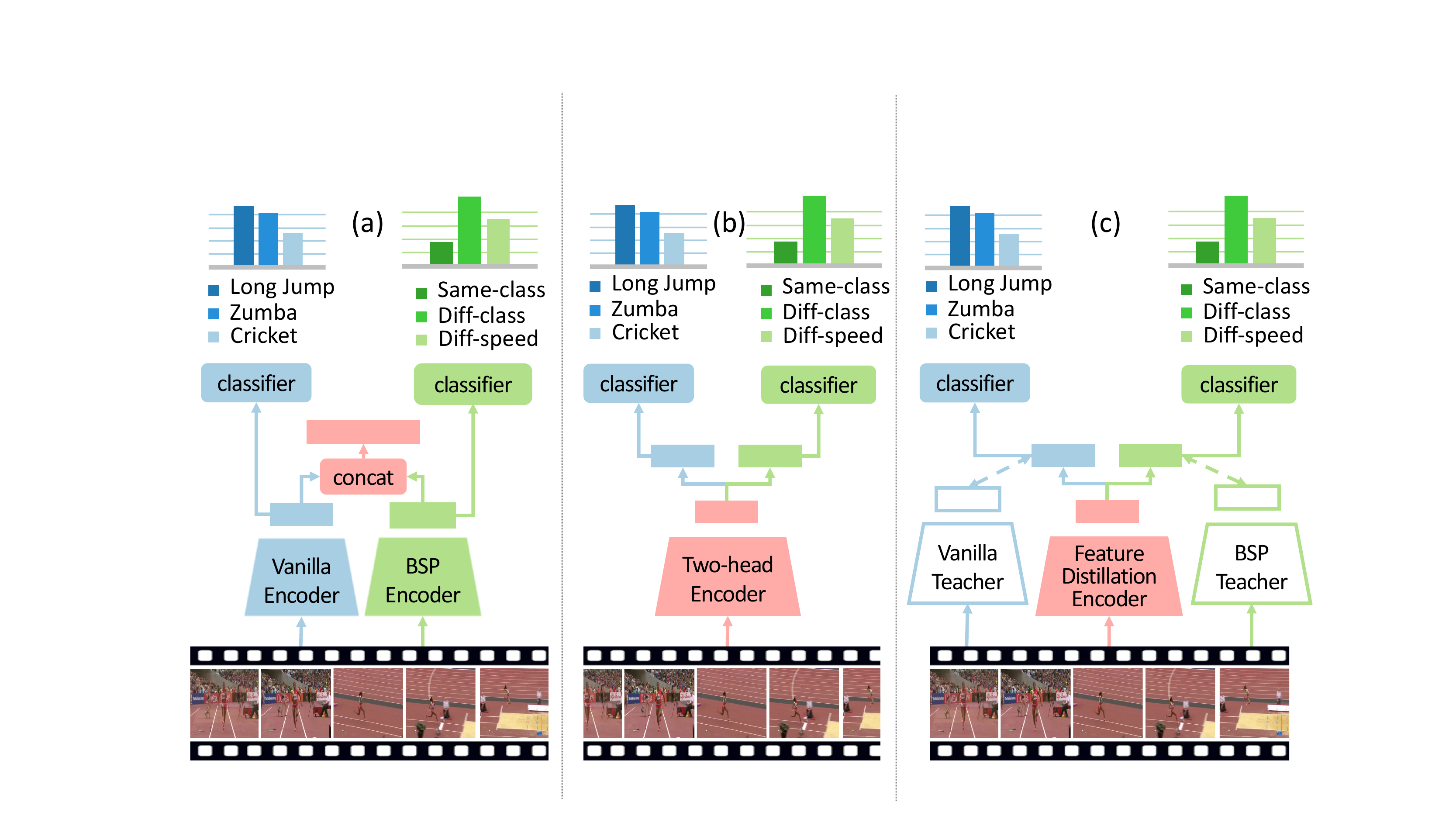}
    \small
    \caption{
    \textbf{Integrating BSP with vanilla action-classification pre-training.}
    (a) Two independently-trained feature streams produce vanilla action and BSP features that are concatenated as output.
    (b) Single feature stream with two-head classification.
    (c) Task specialisation enforced by feature distillation from two teacher streams pre-trained on vanilla and BSP tasks.
    } 
    \label{fig:fuse}
    \vspace{-4mm}
\end{figure}

\subsection{Integration with action classification-based pre-training}
\label{ssec:integration}

We integrate our method with the classification-based pre-training features for enhancing boundary awareness as required for 
temporal localization downstream tasks.
Three architecture designs are considered:
two-steam, two-head, and feature distillation.

{\bf Two-stream.}
This design consists of two steams in parallel,
one for action classification-based pre-training and one for our boundary-sensitive pre-training (Fig.~\ref{fig:fuse}(a)).
For simplicity, we use the same backbone for both.
To integrate their information, feature concatenation 
is adopted at the penultimate layer.

{\bf Two-head.}
In contrast to the two-stream design,
this is a more compact and efficient architecture
with all layers shared for both tasks except the classification layer (Fig.~ \ref{fig:fuse}(b)).
One implicit assumption is that
the two types of feature representation can be 
well fused throughout the feature backbone through end-to-end joint training.

{\bf Feature distillation.}
An alternative approach to the two-stream or two-head network design is to employ a single network, and to train it to produce the same features as the independent networks through the imposition of feature matching losses (Fig.~\ref{fig:fuse}(c)). In particular, let $f_{v}$ be the network trained through standard supervision on Kinetics, and $f_{b}$ a network trained in the proposed self-supervised manner. We then train a single network $f_s$ and two pointwise projection layers, $h_1$ and $h_2$ to minimize a feature matching loss in the form of $\|f_{v}(x)-h_1(f_s(x))\|^2_2 + \|f_{b}(x)-h_2(f_s(x))\|^2_2$, shown as the dashed line in Fig~\ref{fig:fuse}(c). In practice, we also use a standard cross-entropy loss on top of the two feature matching losses.

\section{Experiments}

\subsection{Experiment setup}

\paragraph{Temporal localization tasks. }
In our evaluation, three representative temporal localization tasks for untrimmed videos are considered: temporal action localization~\cite{caba2015activitynet}, video grounding~\cite{Hendricks_2017_ICCV,gao2017tall}, and step localization~\cite{Zhukov2020}.




As discussed in Sec.~\ref{sec:related}, these tasks share the same target of recognizing the specific point in time where the semantic content of the video changes. Solutions to the three tasks use the same two-step training paradigm:
first pre-training the video encoder with the BSP method, 
followed by training the task-specific model
with our BSP model {\em frozen} as video feature extractor.
This allows to explicitly examine the quality and efficacy of model pre-training.

\paragraph{Datasets. }
We use three different video datasets to 
evaluate the performance of temporal localization tasks.

\noindent {\em (1) ActivityNet-1.3}~\cite{caba2015activitynet}  
is a popular benchmark for temporal action localization.
%
It contains 19,994 annotated untrimmed videos with 200 different action classes.
The split ratio of train:val:test
is 2:1:1. 
Each video has an average of 1.65 action instances. Following the common practice, we train and test the models on the training and validation set.

\noindent{\em (2) Charades-STA}~\cite{gao2017tall} 
is a commonly used video grounding dataset extended from 
the action recognition dataset Charades~\cite{sigurdsson2016hollywood}.
It contains $9,848$ videos of daily indoors activities, with $12,408$/$3,720$ moment-sentence pairs in train/test set respectively. 

\noindent{\em (3) CrossTask}~\cite{Zhukov2019}
is an instructional video dataset
depicting complex tasks \eg~make pancakes. We evaluate the step localization performance in terms of actionness prediction~\cite{Zhukov2020}. Following the evaluation protocol described in~\cite{Zhukov2020}, we focus on the 18 primary tasks with temporal annotations \ie~2750 videos with a 3:1 train:test split ratio.


\paragraph{Evaluation metrics. }
We adopt the standard performance metrics specific for each downstream task.
For {\em temporal action localization}, 
mean Average Precision (mAP) at varying temporal Intersection over Union (tIoU) thresholds
is used. 
Following the official evaluation setting, we report mAP scores at three tIoU thresholds of $\{0.5, 0.75, 0.95\}$ and the average mAP over ten thresholds of $[0.05:0.95]$ with step at 0.05 for ActivityNet-1.3. 
For {\em video grounding}, we report the top-1 recall at three different $tIoU$ thresholds, $\{0.3, 0.5, 0.7\}$ for Charades-STA dataset. We also follow LGI's evaluation metrics~\cite{Mun_2020_CVPR} to report the mean tIoU between predictions and ground-truth. 
For \textit{step localization}, we evaluate it in terms of actionness prediction~\cite{Zhukov2020}. We follow its evaluation protocol, dividing the test videos onto non-overlapping  0.2s segments and report the framewise binary average precision (AP). Any segment associated with a step annotation is considered as foreground (positive), and background (negative) otherwise.

\begin{table}[t]
\centering
\caption{\textbf{TAL on ActivityNet-1.3 validation set}. ``*" indicates RGB-only Kinetics pre-trained TSM feature without fine-tuning. 
}
\small
\vskip 0.1cm
\begin{tabular}{p{2.9cm}p{0.62cm}<{\centering}p{0.62cm}<{\centering}p{0.62cm}<{\centering}>{\columncolor[gray]{0.8}}p{0.9cm}<{\centering}}
\toprule
Method  & 0.5  &  0.75  & 0.95 & Average\\
\hline
Singh \textit{et al.} \cite{singh2016untrimmed} & 34.47 & - & - & - \\
Wang \textit{et al.} \cite{wang2016uts}    & 43.65 & -  & - & -\\
Chao \textit{et al.} \cite{chao2018rethinking} & 38.23 & 18.30 & 1.30 & 20.22 \\
SCC \cite{heilbron2017scc}   & 40.00 & 17.90  & 4.70   & 21.70  \\
CDC \cite{shou2017cdc} & 45.30 & 26.00 & 0.20 & 23.80 \\
R-C3D \cite{xu2017r} & 26.80 & - & - & - \\
BSN \cite{lin2018bsn} & 46.45  & 29.96 & 8.02  & 30.03  \\
P-GCN \cite {Zeng2019GraphCN} &48.26 &33.16 &3.27 &31.11  \\
BMN \cite{lin2019bmn} & { 50.07} & {34.78} & { 8.29} & { 33.85}  \\
{BC-GNN}~\cite{Bai2020bcgnn} & {50.56} & {34.75} & {\textbf{9.37}} & { 34.26} \\
\hline
{G-TAD}~\cite{xu2020g} & {50.36} & {34.60} & {{9.02}} & { 34.09} \\
{G-TAD}* & {50.01} & {35.07} & {8.02} & { 34.26} \\
{G-TAD}*+\bf BSP (Ours)  & \textbf{50.94} & \textbf{{35.61}} & {7.98} & {\bf 34.75} \\
\bottomrule
\end{tabular}
\label{tab:sota_anet}
\end{table}

\paragraph{Implementation details. }
Throughout the experiments, we only use RGB input to compute the video representation since optical flow is computationally expensive and adds complexity to the feature extraction model. However, current standard features rely on TSN~\cite{wang2016temporal}, which is insensitive to time. Thus, removing the optical flow stream can have a very negative impact. In order to alleviate this issue, we adopt the Temporal Shift Module (TSM) architecture~\cite{lin2019tsm}.

Given a variable-length video, we firstly sample every 8 consecutive frames as a snippet. Then we feed the snippet into our pre-trained models, and save the features before the fully connected layer. Thus, we obtain a set of snippet-level feature for the untrimmed video.
For {\em language representation} as required in video grounding task,
the pre-processing for text queries includes lowercase conversion and tokenization.
A pre-trained GloVe model~\cite{pennington2014glove} us then used to obtain the initial 300-dimensional word embeddings. A three-layer LSTM~\cite{HochSchm97} follows to produce the feature representation of the input sentence.

For the state-of-the-art {\em temporal localization models},
we select G-TAD~\cite{xu2020g} for temporal action localization on Activity-1.3,
2D-TAN~\cite{2DTAN_2020_AAAI} and LGI~\cite{Mun_2020_CVPR} for video grounding on Charades-STA. A simple linear classifier~\cite{zhu2020vision} is used for step localization in CrossTask.
%

\subsection{Comparison to the state-of-the-art}
In this section, we compare the performance of the proposed BSP features under different tasks and different temporal localization networks. We combine the BSP features with those obtained using classification-based pre-training in a two-stream manner as per Sec.~\ref{ssec:integration}.

On the TAL task, our BSP feature can significantly boost the performance of G-TAD, see Tab.~\ref{tab:sota_anet}. 
Adding BSP features increases performance by 0.93\% at 0.5 IoU, and by 0.5\% at average mAP. Compared to the features originally used by G-TAD, our method \textbf{does not} require time-consuming computations to extract optical flow, \textbf{neither} fine-tune our video encoder on ActivityNet. 
To further validate our method, we conducted experiments on THUMOS-14 and HACS-1.1 dataset against an RGB-only baseline. We observed a gain of $+9.66\%$ mAP@IoU=0.5 and $+0.78\%$ on AmAP respectively. \textbf{Please refer to our supplementary for details}.

We also validate our BSP feature on the video grounding task in Tab.~\ref{tab:sota_charades}. Our BSP benefits both an anchor-based method such as 2D-TAN~\cite{2DTAN_2020_AAAI}  and an anchor-free method such as LGI~\cite{Mun_2020_CVPR}. The latter is also comparable to state-of-the-art performance on Charades-STA.  LGI follows \cite{Rodriguez_2020_WACV} to use a pre-trained video encoder on the downstream dataset. For a fair comparison to BSP, we only include methods that do not fine-tune on Charade-STA.

\begin{table}[t]
\centering
\caption{\textbf{Video grounding on Charades-STA}. ``*" indicates RGB-only Kinetics pre-trained TSM feature without fine-tuning. We use the original evaluation code of LGI and 2D-TAN. 2D-TAN does not compute R@0.3 and mIoU, hence why it is not reported.
}
\small
\begin{tabular}{lcccc} 
\toprule
Method & R@0.3 & R@0.5 & R@0.7 & mIoU \\
\midrule
CTRL~\cite{gao2017tall}    & -     & 21.42 & 7.15 &  - \\
SMRL~\cite{Wang_2019_CVPR}    & -     & 24.36 & 9.01 &  -\\
SAP~\cite{Chen_19_SAP}     & -     & 27.42 & 13.36 &  -\\
MLVI~\cite{xu2019multilevel}    & 54.70 & 35.60 & 15.80 &  -\\
MAN~\cite{zhang2019man} & - & 46.53 & 22.72 & - \\
DRN \cite{Zeng_2020_CVPR} & - & 53.09	& \textbf{31.75}	& - \\\hline
2D-TAN \cite{2DTAN_2020_AAAI} & -   &   42.80 &	23.25    &  -    \\
2D-TAN* & -   &   48.36 &	27.88    &  -    \\
2D-TAN*+{\bf BSP} &   -  &  51.64   &  29.57   &   -  \\\hline
LGI* \cite{Mun_2020_CVPR}    & 60.67   & 45.65   &  23.87   & 43.40  \\
LGI*+{\bf BSP} &   \textbf{68.76}  &  \textbf{53.63}   &   {29.27}   &    \textbf{50.55}  \\
\bottomrule
\end{tabular}
\label{tab:sota_charades}
\end{table}

Table~\ref{tab:sota_crosstask} showcases the results for step localization in terms of actionness prediction. Our baseline, Linear clf., solely pre-trained on Kinetics400, achieves competitive results \wrt~Zhukov~\etal~\cite{Zhukov2020} (47.6\%~\vs46.9\%). Enhancing the Linear clf. with our BSP feature boosts the performance by 1.2\%). These results demonstrate the versatility of our pre-training to capture semantic temporal changes for a different visual domain, namely instructional videos.

\begin{table}[t]
\centering
\caption{
\textbf{Step localization results on CrossTask dataset.} Model performance is shown as framewise average precision (AP) for actionness prediction, as in~\cite{Zhukov2020}.
}
\small
\begin{tabular}{p{2.4cm}p{1.4cm}<{\centering}p{1.7cm}<{\centering}p{0.4cm}<{\centering}}
\toprule
Method & HowTo100M Pretrained & Actionness Supervision  & AP \\
\midrule
Linear clf.~\cite{Zhukov2020} & \cmark & \cmark & 56.2 \\
Zhukov~\etal~\cite{Zhukov2020} & \cmark & \xmark & 47.6 \\
\midrule
Linear clf.  & \xmark & \cmark & 46.9 \\
Linear clf. + \textbf{BSP}  & \xmark & \cmark & 48.1 \\
\bottomrule
\end{tabular}
\label{tab:sota_crosstask}
\end{table}

\begin{figure}[t]
    \centering
    \includegraphics[trim={8.5cm 2cm 7.5cm 1cm},width=7.5cm,clip]{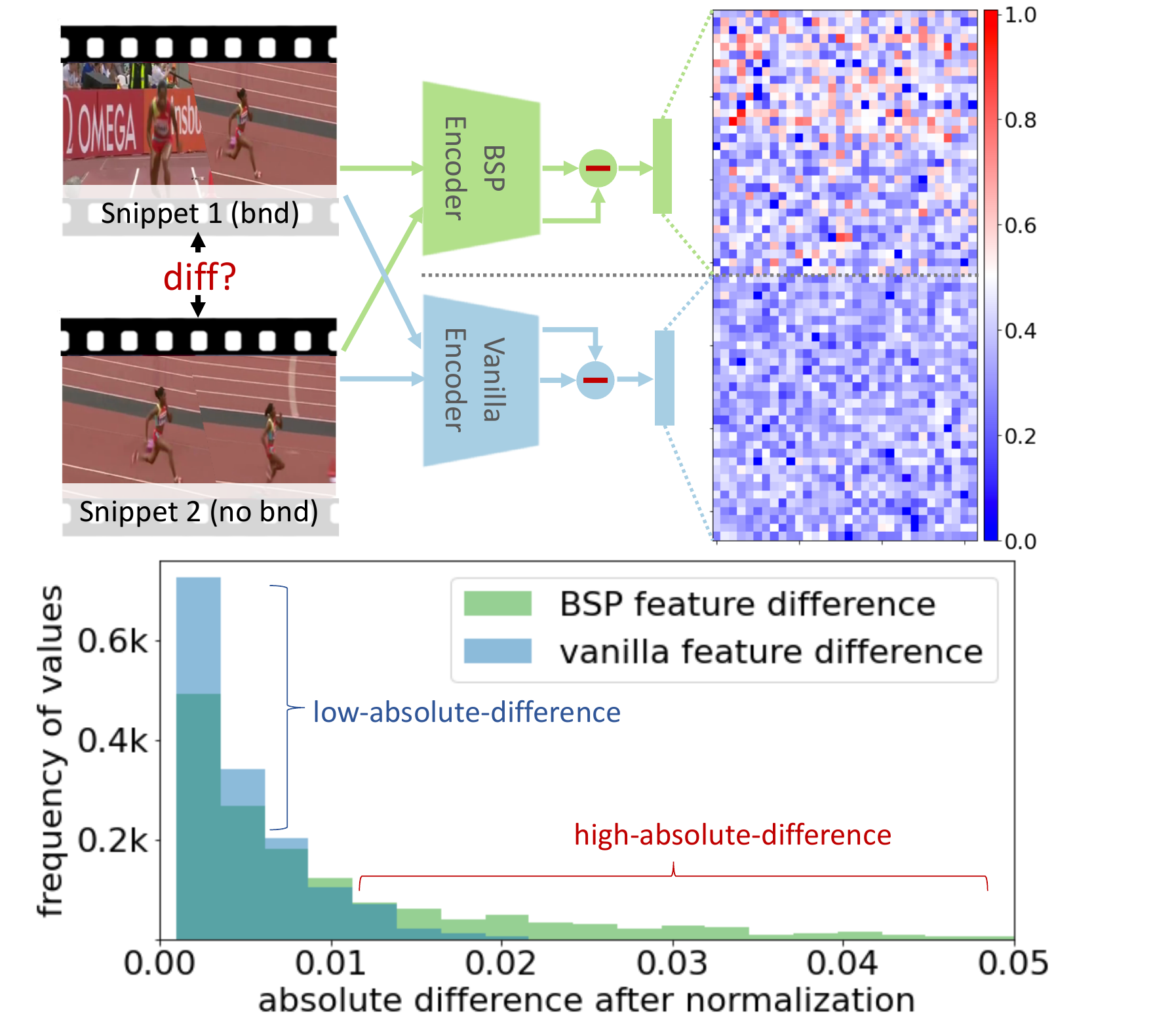}
    \caption{ \textbf{Visualization of BSP snippet-feature and vanilla snippet-feature.} 
    \textit{Top}: we visualize the consecutive snippet feature difference from BSP and vanilla video encoders. 
    Comparing the absolute differences, our BSP encoder produces distinguishable snippet representations sensitive to the boundary.
    \textit{Bottom}: The distributions of the absolute difference of boundary and non-boundary snippets from BSP and vanilla encoders show that the vanilla encoder fails to properly capture boundary information. 
    } 
    \label{fig:vis}
    \vspace{-2mm}
\end{figure}

\subsection{Feature Visualization}

We compare the BSP and vanilla snippet-feature in Fig.~\ref{fig:vis}. 
Concretely, we visualize the absolute difference of feature representations from two consecutive snippets, while a natural boundary only appears in the first snippet. 
The feature representations are reshaped into $32\times 32$ matrices, and the absolute differences are visualized in the top-right of Fig.~\ref{fig:vis}. The upper matrix from BSP video encoder contains more high-difference (red) values, indicating that our method produces distinguishable snippet representations that are sensitive to boundary.
Moreover, we compute the distribution of the absolute difference of the two snippets from BSP and vanilla encoders, and visualize them in the bottom of Fig.~\ref{fig:vis}. From the histogram plot, we can have the same observation that the vanilla encoder lacks the capability of capturing boundary information. 
For better comparison and visualization, the feature vectors are normalized and reshaped from $2048$ to $64\times32$. 
    
\subsection{Ablation studies}
\label{sec:abl}

To verify the effectiveness of our proposed boundary-sensitive pre-training strategy, we perform in-depth ablation analysis of different boundary-sensitive tasks, fusion methods, and backbone depth. For the ease of experimentation, all the ablation studies employ a ResNet18 backbone.

\paragraph{Boundary-sensitive tasks}
We first show in Tab.~\ref{tab:abl1} the performance results for a number of pretext tasks. Given the four types of augmented data, $S=\{S_{dc},S_{sc},S_{ds},S_{ss}\}$, we consider the regression loss described in Sec.~\ref{ssec:pretexts} for each individual task, binary classification task for each of the pretext tasks, and the final 4-way classification task. We further include a randomly initialized video encoder as a baseline. Although four different boundary classes together ($S$) output the best overall performance, the change of speed ($\{S_{ds},S_{ss}\}$) gives more boost than the change of class semantics, ($\{S_{dc},S_{ss}\}$ and $\{S_{sc},S_{ss}\}$).
Moreover, the classification task is consistently better than the regression counterpart.
We believe this is due to it being a more challenging learning task and its inability to exploit the original source data.
Moreover, for temporal localization, the model needs to understand whether a segment completes, \ie, boundaries are present, rather than where exactly the boundary is.
It is also clear that there is still a significant performance drop against classification-based pre-training. Only leveraging boundary information is not enough to localize actions, as global semantic information is also required. This is given by classification-based pre-training.

\begin{table}[t]
\centering
\caption{\textbf{Different boundary-sensitive pretext tasks}. We compare our four pretext tasks with two loss functions. 
The random baseline refers to
a randomly initialized video encoder, while the vanilla baseline refers to the video encoder pre-trained on Kinetics-400 in a fully supervised
manner. 
We compare the performance of G-TAD on ActivityNet 1.3 dataset. 
}
\vspace{0.2cm}
\small
\begin{tabular}{p{1.3cm}<{\centering}p{0.5cm}<{\centering}p{0.62cm}<{\centering}p{0.62cm}<{\centering}p{0.62cm}<{\centering}>{\columncolor[gray]{0.8}}p{0.9cm}<{\centering}}
\toprule
Edition & Task  & 0.5  &  0.75  & 0.95 & Average\\
\midrule
\multicolumn{2}{l}{random-baseline} & {40.44} & {24.99} & {6.85} & {25.58} \\ \hline
$\{S_{dc}\}$  & reg & 40.88 & 26.64 & 5.73 & 26.56 \\ 
$\{S_{sc}\}$ & reg & 42.48 & 27.78 & 6.56 & 27.75 \\
$\{S_{ds}\}$   & reg & 44.00 & 29.05 & 4.82 & 28.77 \\
$\{S_{dc},S_{ss}\}$  & cls & 44.32 & 29.16 & \textbf{6.66} & 29.17 \\
$\{S_{sc},S_{ss}\}$ & cls & 44.24 & 29.20 & 6.25 & 29.09 \\
$\{S_{ds},S_{ss}\}$   & cls & 45.00 & 29.89 & 6.47 & 29.75 \\
$S$   & cls & \textbf{45.39} & \textbf{30.26} & 6.33 & \textbf{29.97} \\ \hline
\multicolumn{2}{l}{vanilla-baseline} & 49.64 & 34.16 & 7.68 & 33.59 \\
\bottomrule
\end{tabular}
\label{tab:abl1}
\end{table}

\begin{table}[!ht]
\centering
\caption{\textbf{Different feature fusion methods.} We compare three fusion methods: two-stream, two-head, and feature distillation. The
performance is measured by Recall (\%) of 2D-TAN on Charades and mAP (\%) of G-TAD on ActivityNet
1.3 dataset.
}
\vspace{0.2cm}
\small
\begin{tabular}{p{1.5cm}<{\centering}|p{0.6cm}<{\centering}p{0.6cm}<{\centering}|p{0.6cm}<{\centering}p{0.6cm}<{\centering}p{0.6cm}<{\centering}p{0.6cm}<{\centering}}
\toprule
Methods &   \multicolumn{2}{c|}{2D-TAN}&  \multicolumn{4}{c}{G-TAD} \\  \hline
tIoU & 0.5 & 0.7  & 0.5  &  0.75  & 0.95 & Avg \\ 
\midrule
{vanilla} & 39.78 & 22.15 & 49.64 & 34.16 & 7.68 & 33.59  \\ \hline
 2-stream & \underline{44.01} & \textbf{24.95} & \textbf{50.09} &\textbf{34.66} & \underline{7.95} & \textbf{33.96}\\
 2-head & 39.95 & 22.85  & 49.50 & 34.00 & \textbf{8.38} & 33.54 \\
 feat dist & \textbf{44.65} & \underline{24.73} &  \underline{49.67} & \underline{34.40} & 7.74 & \underline{33.74}\\
\bottomrule
\end{tabular}
\label{tab:abl2}
\end{table}

\paragraph{Feature Integration}
We further investigate different ways to integrate both features, including two-stream, two-head, and feature distillation. We conduct experiments on both G-TAD and 2D-TAN. According to Tab.~\ref{tab:abl2}, the two-stream method gives the best overall performance for temporal localization tasks. Among the single-network solutions, feature distillation is better than multi-task learning (two-head).



\paragraph{Robustness to model backbone and capacity}
We investigate the impact of different model backbones and capacity in~Tab.~\ref{tab:18vs50}. We use G-TAD for the TAL task on ActivityNet 1.3, with TSM-18, TSM-50, and R(2+1)D-34 as backbone choices. It is clear that a deeper model produces better BSP features, and that the net contribution when concatenated to classification-based BSP features is consistently and similarly positive in all cases. 
This verifies the general efficacy of our method on different backbones with varying capacity.

\begin{table}[t]
\centering
\caption{\textbf{Model backbone and capacity.} We compare performance when using TSM-18, TSM-50, and R(2+1)d-34 for G-TAD on ActivityNet 1.3 dataset. }
\vspace{0.2cm}
\small
\begin{tabular}{p{1.5cm}p{1cm}<{\centering}p{0.62cm}<{\centering}p{0.62cm}<{\centering}p{0.62cm}<{\centering}>{\columncolor[gray]{0.8}}p{0.9cm}<{\centering}}
\toprule
Backbone & BSP   & 0.5  &  0.75  & 0.95 & Average\\
\midrule
TSM-18  & \xmark & 49.64 & 34.16 & 7.68 & 33.59 \\
TSM-18  & \cmark & {50.09} &{34.66} & 7.95 & {33.96} \\ \hline
TSM-50  & \xmark & 50.32 & 35.07 & 8.02 & 34.26 \\
TSM-50  & \cmark & {50.94} &{35.61} & {7.98} & {34.75} \\ \hline
R(2+1)d-34  & \xmark & 49.57 & 34.92 & 8.43 & 34.05 \\
R(2+1)d-34  & \cmark & 50.28 & 35.65 & 8.06 & 34.53 \\

\bottomrule
\end{tabular}
\label{tab:18vs50}
\end{table}


\section{Conclusion}
In this work we have investigated the under-studied problem of model pre-training for temporal localization tasks in videos
by introducing a novel Boundary-Sensitive Pretext (BSP) task.
Beyond vanilla pre-training on trimmed video data, we exploited a large number of videos with different types of synthetic temporal boundary information,
and explored a number of pretext task designs using these
boundary-sensitive videos.
We evaluated extensively the BSP model
on three representative temporal localization tasks with different input modalities and motion complexity.
The results demonstrate that our BSP can strongly enhance the vanilla model with boundary-sensitive feature representations, yielding competitive or new state-of-the-art performance
on these temporal localization tasks.




\renewcommand\thesection{\Alph{section}}
\setcounter{section}{0}

\section{Comparison on other TAL datasets}
To further validate our method, we conduct experiments on two more datasets with different sizes for temporal action localization. On both datasets, we firstly use a Kinetics pre-trained TSM-50 model to extract feature and then run G-TAD on top of it. Then, for fair comparison, we use our BSP feature extractor for the same process. Both experiments are tested with the same environment and same hyper-parameters. 

\subsection{Evaluation on larger dataset}
Human Action Clips and Segments ({\bf HACS-1.1})
\cite{zhao2019hacs}
is a recent large-scale temporal action localization dataset.
It contains 140K complete segments on 50K videos in 200 action categories. Compared with ActivityNet holding 20K videos, HACS dataset is larger and more challenging.

We show the performance of BSP on HACS-1.1 dataset against an RGB-only baseline in Tab.~\ref{tab:h}. We observe a gain of $+0.78\%$ on Average mAP, shown by the grey column of the table.

\begin{table}[h]
\centering
\caption{\textbf{Evaluations on HACS-1.1 dataset on temporal action localization.} ``*" indicates RGB-only Kinetics pre-trained TSM feature without fine-tuning.
}
\vspace{0.2cm}
\small
\begin{tabular}{p{2.2cm}p{0.62cm}<{\centering}p{0.62cm}<{\centering}p{0.62cm}<{\centering}>{\columncolor[gray]{0.8}}p{0.9cm}<{\centering}}
\toprule
Method  & 0.5  &  0.75  & 0.95 & Average\\
\midrule
G-TAD* & 37.50 & 23.80 & 7.10 & 24.25 \\ \hline
G-TAD* +BSP&  38.10 & 24.73 & 7.76 & 25.03 \\ \hline
\textit{Gain} & +0.60 & +0.93 & +0.66 & +0.78 \\
\bottomrule
\end{tabular}\label{tab:h}
\end{table}

\subsection{Evaluation on smaller dataset}
{\bf THUMOS-14} \cite{jiang2014thumos} dataset contains 413 temporally annotated untrimmed videos with 20 action categories. We use the 200 videos in the validation set for training and evaluate on the 213 videos in the testing set. Different from ActivityNet, THUMOS-14 has in average 16 action instances per video for the test set. Thus, the sensitivity to action boundaries plays a more important role for its temporal action localization. 

The performance of our method on THUMOS-14 dataset against an RGB-only baseline is shown in Tab.~\ref{tab:h}. As shown by the grey column of the table, a significant gain of $+9.66\%$ on mAP at IoU=0.5 further demonstrates the effectiveness of our proposed approach.

\begin{table}[t]
\centering
\caption{\textbf{Evaluation of temporal action localization on THUMOS-14 dataset.} ``*" indicates RGB-only Kinetics pre-trained TSM feature without fine-tuning.
}
\vspace{0.2cm}
\small
\begin{tabular}{p{2.2cm}p{0.62cm}<{\centering}p{0.62cm}<{\centering}>{\columncolor[gray]{0.8}}p{0.62cm}<{\centering}p{0.62cm}<{\centering}p{0.62cm}<{\centering}}
\toprule
Method  & 0.3  &  0.4  & 0.5 & 0.6 & 0.7\\
\midrule
G-TAD* & 46.61 & 39.46 & 30.14 & 20.13 & 12.15 \\ \hline
G-TAD* +BSP&  52.34 & 46.28 &39.80 & 30.81 & 21.14 \\ \hline
\textit{Gain} & +5.73 & +6.82 & +9.66 & +10.68 & +8.99 \\
\bottomrule
\end{tabular}
\label{tab:abl1}
\end{table}

\section{Comparison on other TAL methods}
To test the robustness of BSP to different TAL methods, we also conduct  experiments on Boundary-Matching Network (BMN)~\cite{lin2019bmn} based on a publicly available re-implementation\footnote{https://github.com/JJBOY/BMN-Boundary-Matching-Network}.
BMN is an effective, efficient and end-to-end trainable proposal generation method, which generates proposals with precise temporal boundaries and reliable confidence scores simultaneously. 
BMN proposed the Boundary-Matching (BM) mechanism to evaluate confidence scores of densely distributed temporal action proposals.
Please note that BMN targets the problem of temporal action proposal generation, which is a sub-task of temporal action localization. In the proposal generation task, models only predict actions segments, but do not assign action labels to them. Thus, the evaluation metric of BMN is different than for TAL. The proposal generation task is evaluated by the top-k average recall of predictions (AR@k), where $k\in \{1,5,10,100\}$. The area under the recall curve (AUC) is used to evaluate the overall performance.

It can be observed from Tab.~\ref{tab:bmn} that BSP brings consistent improvements over all the evaluation metrics on BMN, showing that our method can generalize to different methods. Particularly, our method can get 67.61 AUC which is close to the value, 67.7, reported by the code repository. However, no optical flow is used in BSP method.
\begin{table}[h]
\centering
\caption{\textbf{Evaluation of temporal action proposal generation on ActivityNet.} ``*" indicates RGB-only Kinetics pre-trained TSM feature without fine-tuning.
}
\vspace{0.2cm}
\small
\begin{tabular}{p{1.7cm}p{0.75cm}<{\centering}p{0.75cm}<{\centering}p{0.80cm}<{\centering}p{1.0cm}<{\centering}>{\columncolor[gray]{0.8}}p{0.75cm}<{\centering}}
\toprule
Method  & AR@1 & AR@5 & AR@10 & AR@100 & AUC\\
\midrule
BMN* &  33.45 &         49.41 & 56.55 & 75.17 & 67.26\\ \hline
BMN* +BSP&  33.69 &     50.12 & 57.35 & 75.50 & 67.61\\ \hline
\textit{Gain} & +0.24 & +0.71 & +0.80 & +0.33 & +0.35\\
\bottomrule
\end{tabular}
\label{tab:bmn}
\end{table}

\section{Comparison to other pre-training methods}

We compare BSP to two other popular video self-supervised methods in Tab.~\ref{tab:ssl}: Arrow of Time~\cite{arrow_cvpr18} and SpeedNet~\cite{benaim2020speednet}. Furthermore, we also include the results when ensembling two classification-based pre-trained models and for a randomly-initialized network. All comparisons use the two-steam integration with the same classification-based pre-trained TSM-18 video encoder. 
A random model does not give extra information, while 
Speednet~\cite{benaim2020speednet} and Arrow of Time~\cite{arrow_cvpr18} can enrich the original features. Since they are not sensitive to the action boundaries, these methods are experimentally equivalent to training an ensemble of two vanilla video encoder (last row). Their performance is thus clearly inferior to that of our BSP method.


\begin{table}[h]
\centering
\caption{\textbf{Comparison to other pre-training methods.} We compare BSP to other pre-training strategies. The results show TAL performance of G-TAD on THUMOS-14 dataset. 
}
\vspace{0.2cm}
\small
\begin{tabular}{p{1.5cm}p{0.62cm}<{\centering}p{0.62cm}<{\centering}>{\columncolor[gray]{0.8}}p{0.62cm}<{\centering}p{0.62cm}<{\centering}p{0.62cm}<{\centering}}
\toprule
Method   & 0.3  &  0.4  & 0.5 & 0.6 & 0.7\\
\midrule
baseline   & 41.48 & 34.21 & 25.88 & 17.82 & 11.28 \\ \hline
Arrow \cite{arrow_cvpr18}  & 46.52 & 40.13 & 31.30 & 22.24 & 14.39  \\
Speed \cite{benaim2020speednet}  & 46.76 & 39.49 & 31.92 & 23.21 & 15.18  \\
\bf BSP (ours)   & \textbf{48.71} & \textbf{42.78} & \textbf{34.92} & \textbf{27.44} & \textbf{17.60} \\ \hline
ensemble  & 45.67 & 37.70 & 28.66 & 19.55 & 11.32 \\
\bottomrule
\end{tabular}
\label{tab:ssl}
\end{table}

{\small
\bibliographystyle{ieee_fullname}
\bibliography{egbib}
}

\end{document}